\documentclass{article}

\usepackage{arxiv}

\usepackage{amsmath,amsfonts}
\usepackage{algorithmic}
\usepackage{algorithm}
\usepackage{array}
\usepackage[caption=false,font=normalsize,labelfont=sf,textfont=sf]{subfig}
\usepackage{textcomp}
\usepackage{stfloats}
\usepackage{url}
\usepackage{verbatim}
\usepackage{graphicx}
\usepackage{cite}

\usepackage{mathtools}
\usepackage{amsfonts}
\usepackage{amsmath}
\usepackage{amssymb}
\usepackage{bm}

\usepackage{dsfont}

\usepackage{comment}

\DeclareMathOperator{\diag}{diag}
\DeclareGraphicsExtensions{.png,.pdf}

\begin{document}
\title{Accelerated sparse Kernel Spectral Clustering for large scale data clustering problems}

\author{
            Mihaly Novak\\
            Software for Experiments group of the Experimental Physics Department\\
            CERN\\ 
            1211 Geneva 23 - Switzerland\\
            \texttt{mihaly.novak@gmail.com}\\
            \AND
            Rocco Langone\\
            Department of Electrical Engineering (ESAT-STADIUS)\\
            Katholieke Universiteit Leuven\\ 
            Leuven B-3001, Belgium\\
            \texttt{roccolangone@hotmail.com}\\
            \AND
            Carlos Alzate\\
            AI Fund\\
            Palo Alto, CA 94306, USA\\
            \texttt{carlos.alzate@gmail.com}\\
            \AND
            Johan Suykens\\
            Department of Electrical Engineering (ESAT-STADIUS)\\
            Katholieke Universiteit Leuven\\ 
            Leuven B-3001, Belgium\\
            \texttt{johan.suykens@esat.kuleuven.be}
}

\date{}

\renewcommand{\headeright}{ }
\renewcommand{\undertitle}{ }

\maketitle
\thispagestyle{empty}
\begin{abstract}
An improved version of the sparse multiway kernel spectral clustering (KSC) is presented in this brief. 
The original algorithm is derived from weighted kernel principal component (KPCA) analysis formulated within the primal-dual
least-squares support vector machine (LS-SVM) framework. Sparsity is achieved then by
the combination of the incomplete Cholesky decomposition (ICD) based low rank approximation
of the kernel matrix with the so called reduced set method. The original ICD based sparse KSC 
algorithm was reported to be computationally far too demanding, especially when applied on large scale data 
clustering problems that actually it was designed for, which has prevented to gain more than simply theoretical 
relevance so far. This is altered by the modifications reported in this brief that drastically improve the computational characteristics.
Solving the alternative, symmetrized version of the computationally most demanding core eigenvalue problem eliminates the necessity 
of forming and SVD of large matrices during the model construction. This results in solving clustering problems now within seconds that were 
reported to require hours without altering the results. Furthermore, sparsity is also improved significantly, leading to more compact 
model representation, increasing further not only the computational efficiency but also the descriptive power. These transform the original, 
only theoretically relevant ICD based sparse KSC algorithm applicable for large scale practical clustering problems. 
Theoretical results and improvements are demonstrated by computational experiments on carefully selected 
synthetic data as well as on real life problems such as image segmentation.
\end{abstract}



\section{Introduction}
\noindent
Spectral Clustering (SC) algorithms are known to perform well even in case of complex structures in the data when classical methods fail. This has led to several
successful applications in computer vision \cite{Shi&Malik_Ncut,zhang2008spectral,arbelaez2010contour}, load balancing \cite{ding2012two,alzate2013improved},
bioinformatics \cite{snel2002identification,higham2007spectral}, network \cite{lei2015consistency,liu2018global} and many other data analysis and machine learning related problems.

Classical formulations of SC \cite{Shi&Malik_Ncut,Chung_spctral_Graph_th,NJW_alg,Hagen&Kahng_Ratiocut,meilua2001random,Luxburg_tut_on_ksc}
starts from a graph partitioning problem that is NP-hard due to the discrete constraints on the indicators. By letting the indicator vectors to be real valued, the optimal solution
of this relaxed version is provided by some of the eigenvectors of the underlying normalised affinity matrix. The discrete cluster indicators of the original problem can then be inferred
from these eigenvectors.

Traditional SC algorithms are typically based on the complete data set due to the lack of a clear extension to unseen data. Since the memory requirement of the affinity matrix
as well as the computing time of the corresponding eigenvalue decomposition grow quickly with increasing data size, these SC algorithms are suitable only for relatively small
size problems. A possible extension is offered in \cite{Bengio_outofsample,Fowlkes_Nystrom} by using the Nystr\"{o}m method \cite{Nytrom_original}. These algorithms rely only on a subset of the data
to obtain an approximation of the implicit eigenfunctions that is used then to cluster the out-of-sample data points.
However, the underlying clustering model is unknown and the hyper-parameter selection is done in a heuristic way.

Unlike the traditional graph partition based SC, the so called Kernel Spectral Clustering (KSC)\cite{alzate&suykens2006_originalwkpca,alzate&suykens2010} has been formulated as weighted Kernel 
Principal Component Analysis (KPCA)\cite{scholkopf1998nonlinear} using the primal-dual Least Square Support Vector Machine (LS-SVM) framework \cite{Johan2003_LS_SVM_kernel_PCA,LS_SVM_book}. 
It has been shown in \cite{alzate&suykens2006_originalwkpca}, that the optimal solution of this weighted KPCA problem in the dual space resembles to the \textit{random walk} SC eigenvalue problem \cite{Shi&Malik_Ncut} 
when choosing the weights appropriately. However, the weight centered kernel matrix plays the role of the affinity matrix in the case of weighted KPCA hence the name KSC. Unlike the classical SC algorithms, 
KSC provides model selection criterion for hyper-parameter tuning that can be used to find the optimal number of clusters and kernel parameters. The most important advantage of KSC though is the natural and straightforward way that it offers for out-of-sample extension without relying any Nystr\"{o}m like approximations. This 
makes possible to construct the clustering model based on a subset of the data while the obtained model can be used then to assign any remaining or new data 
points to the clusters.

For extending the KSC applicability to large scale data, sparse models have been developed \cite{alzate&suykens_KSC_ICD,Alzate&Suykens_sparseKSC_models2011} based on the reduced set method \cite{Burges_simpliSVdecision,Scholkopf_inputVsFeature} and exploiting its out-of-sample extension capability. 
One \cite{Alzate&Suykens_sparseKSC_models2011} relies on an iterative, quadratic R\'{e}nyi entropy 
maximisation based subset selection \cite{Girolami_Renyi_max,LS_SVM_book} before obtaining the reduced set points. A mathematically more concise solution 
\cite{alzate&suykens_KSC_ICD, Alzate&Suykens_sparseKSC_models2011} is built on the Incomplete Cholesky Decomposition (ICD) based low rank approximation 
of the kernel matrix \cite{Fine&Scheinberg_lowrank_aprx2002,Bach&Jordan_ICA2002,Bach&Jordan_lowrank_2005}. 
While both outperform the Nystr\"{o}m approximation based sparse version of the traditional SC \cite{Alzate&Suykens_sparseKSC_models2011}, 
the ICD based algorithm has the great advantage that it automatically provides the reduced set, used for the sparse model construction, while the 
former requires an additional L$_1$+L$_2$ penalisation \cite{Boyd_convexOpt_book,Scholkopf_inputVsFeature} based step for this.
On the other hand, the ICD based sparse KSC was observed \cite{Alzate&Suykens_sparseKSC_models2011} to require larger reduced set size (i.e. worse sparsity) compared to the R\'{e}nyi entropy 
maximisation based version. More importantly, the ICD based sparse KSC algorithm was found to be computationally far too demanding \cite{alzate&suykens_KSC_ICD,Alzate&Suykens_sparseKSC_models2011}, 
especially in case of larger data sets which the sparse algorithm was actually designed for which has prevented to
gain any practical importance so far. Our main contributions, summarized below, result in a new version of 
the algorithm that alter this situation:
\begin{enumerate}
\item{The computationally expensive core part of the original algorithm is replaced with a significantly faster but equivalent alternative.         
         This drastically accelerates the ICD based sparse KSC, especially in case of large data sets, resulting in solving the same clustering 
         problem within seconds that were reported to require hours for the original version without altering the results.}
\item{A more accurate computation of the approximated bias terms, depending less directly on the reduced set size, is introduced. This leads to a 
         substantially increased sparsity of the obtained clustering model which not only improves further the computational efficiency but 
         also results in a more compact model representation.}         
\item{The theoretical results and improvements are demonstrated by computational experiments on carefully selected 
         synthetic as well as on real life problems such as image segmentation.}         
\end{enumerate}



\section{Problem Formulation}
\noindent

\subsection{SC formulation as weighted KPCA}
\label{sec:SC_as_wKPCA}
\noindent
While all the details can be found elsewhere \cite{alzate&suykens2006_originalwkpca, alzate&suykens2010}, the most important characteristics of the (none-sparse) multiway KSC is summarized first briefly in order to provide the necessary basis for the corresponding sparse problem formulation. 

Given an input data set $\mathcal{D}=\{\bm{x}_{i}\}_{i=1}^{N},\bm{x}_{i}\in\mathbb{R}^{d}$
together with the corresponding weights $\mathcal{V}=\{v_{l}\}_{l=1}^{N},v_{l}\in\mathbb{R}^{+}$,
the goal of weighted KPCA is to find the directions such that the weighted variance of the projections of the 
weight centered $\varphi(\cdot):\mathbb{R}^{d} \to \mathbb{R}^{n_{h}}$ feature map of $\mathcal{D}$ onto these $\bm{w}^{(k)} \in \mathbb{R}^{n_{h}}, k=1,\dotsc,\mathcal{K}<N$
direction vectors is maximal. This leads to the following primal optimisation problem \cite{alzate&suykens2006_originalwkpca, alzate&suykens2010}
\begin{equation}
\label{eq:mulitiway_Primal}
\begin{array}{ >{\displaystyle}r >{\displaystyle}c >{\displaystyle}l}
       \max_{\bm{w}^{(k)},\bm{e}^{(k)},b^{(k)}}
	  J(\bm{w},\bm{e},b) & = & \frac{1}{2} \sum_{k=1}^{\mathcal{K}} \gamma^{(k)} \bm{e}^{(k)^{T}}V\bm{e}^{(k)} \\
	    & & -\frac{1}{2}\sum_{k=1}^{\mathcal{K}}\bm{w}^{(k)^{T}}\bm{w}^{(k)}
	    \\ \\
       \mathrm{such\;that} \quad \bm{e}^{(k)} &= &\Phi\bm{w}^{(k)}+b^{(k)}\bm{\mathds{1}}_{N}, \quad k=1,...,\mathcal{K}  \\ \\
\end{array}
\end{equation}
\noindent
where $\gamma^{(k)} \in \mathbb{R}^{+}$, $V\in \mathbb{R}^{N\times N}, [V]_{ii}=v_{i}, i=1,...,N$ is the diagonal weight matrix,
$\Phi = [\varphi(\bm{x}_{1}), \dots, \varphi(\bm{x}_{N})  ]^T \in \mathbb{R}^{N\times n_{h}}$ is the feature map matrix 
and $\bm{e}^{(k)} = \Phi\bm{w}^{(k)}+b^{(k)}\bm{\mathds{1}}_{N} \in \mathbb{R}^{N}, k=1,\dotsc,\mathcal{K}$ are the error vectors 
with the $b^{(k)}\in \mathbb{R}, k=1,\dotsc,\mathcal{K}$ bias terms. It can be shown, that using the bias terms leads to the same weighted 
centering of the feature map as the corresponding explicit centering \cite{Johan2003_LS_SVM_kernel_PCA}.   
The Lagrangian of this constrained optimization problem is
\begin{equation}
\begin{split}
\label{eq:mulitiway_Lagrangian}
  \mathcal{L} &
  (\bm{w}^{(k)},\bm{e}^{(k)},b^{(k)}; \pmb{\beta}^{(k)})
    = \frac{1}{2} \sum_{k=1}^{\mathcal{K}} \gamma^{(k)} \bm{e}^{(k)^{T}}V\bm{e}^{(k)} \\
  & -\frac{1}{2}\sum_{k=1}^{\mathcal{K}} \bm{w}^{(k)^{T}}\bm{w}^{(k)}
    -\sum_{k=1}^{\mathcal{K}} \pmb{\beta}^{(k)^{T}} (\bm{e}^{(k)}
    -\Phi\bm{w}^{(k)}-b^{(k)}\bm{\mathds{1}}_{N} )
\end{split}
\end{equation}
with $\pmb{\beta}^{(k)} \in \mathbb{R}^{N},k=1,...,\mathcal{K}$ Lagrange multiplier vectors.
Starting from the Karush-Kuhn-Tucker (KKT) optimality conditions and eliminating the primal variables 
yield the following eigenvalue problem  
\begin{equation}
\label{eq:multiway_eigenproblem1}
  VM_{v}\Omega\pmb{\beta}^{(k)}=\lambda^{(k)}\pmb{\beta}^{(k)},
    \quad k=1,...,\mathcal{K}
\end{equation}
for the optimal solution of the wighted KPCA problem involving the dual variables $\pmb{\beta}^{(k)}$ as eigenvectors.
$M_{v}=I_{N} -  \bm{\mathds{1}}_{N}\bm{\mathds{1}}_{N}^{T}V / [\bm{\mathds{1}}_{N}^{T}V\bm{\mathds{1}}_{N}]$ above
is the weighted centering matrix, $\Omega=\Phi\Phi^{T}$ is the kernel matrix with $[\Omega]_{ij}=\varphi(\bm{x}_{i})^{T}
\varphi(\bm{x}_{j})=K(\bm{x}_{i},\bm{x}_{j}), i,j=1,...,N$ and $K:\mathbb{R}^{d}\times
\mathbb{R}^{d} \to \mathbb{R}$ is a positive definite kernel while $\lambda^{(k)}=1/\gamma^{(k)}$. It can be shown easily, that the objective given 
by (\ref{eq:mulitiway_Primal}) can be maximised by taking the $\pmb{\beta}^{(k)}, k=1,\dotsc,\mathcal{K}<N$ such that 
$\lambda^{(1)}\geq\lambda^{(2)}\dotsc \geq \lambda^{(\mathcal{K})}$ leading eigenvectors of the $VM_{v}\Omega$ matrix.

The KKT optimality conditions (Eqs.(11) in \cite{alzate&suykens2010}) provide the $\bm{w}^{(k)}=\Phi^{T}\pmb{\beta}^{(k)}$ connection between the primal - dual solutions 
as well as the $b^{(k)}=-\bm{\mathds{1}}_{N}^{T}V\Phi\Phi^{T}\pmb{\beta}^{(k)}  /  [\bm{\mathds{1}}_{N}^{T}V\bm{\mathds{1}}_{N}]$ expression for the bias terms.  
These yield $\bm{z}^{(k)} = \Phi\bm{w}^{(k)}+b^{(k)}\bm{\mathds{1}}_{N} = M_{v}\Omega\pmb{\beta}^{(k)} \in \mathbb{R}^N$ for the $k$-th score variable,  
i.e. projection of the weight centered feature map of the input data set $\mathcal{D}$ onto the $k$-th optimal direction vector.   

As discussed in depth in \cite{alzate&suykens2006_originalwkpca, alzate&suykens2010}, (\ref{eq:multiway_eigenproblem1}) becomes 
\begin{equation}
\label{eq:multiway_eigenproblem}
  D^{-1}M_{D}\Omega\pmb{\beta}^{(k)}=\lambda^{(k)}\pmb{\beta}^{(k)},
    \quad k=1,...,\mathcal{K}
\end{equation}
when using the $K(\bm{x}_{i},\bm{x}_{j})$ kernel to measure the pairwise similarities between the $\bm{x}_{i},\bm{x}_{j} \in \mathcal{D}$ input data points 
and choosing the corresponding inverse degree matrix $V=D^{-1}, \diag(D)=\Omega\bm{\mathds{1}}_{N}$ as the weight matrix. This resembles to the eigenproblem 
of a classical SC algorithm, introduced in \cite{Shi&Malik_Ncut} based on the normalized \textit{random walk} graph Laplacian, with the only difference that now 
the weight centered kernel matrix plays the role of the affinity matrix. However, the properties of the eigenvectors $\pmb{\beta}^{(k)}$ are different now from those 
obtained during the related classical SC algorithm due to the weighted centering of the kernel matrix. Nevertheless, the special properties of some of these $\pmb{\beta}^{(k)}$ 
eigenvectors make possible the clustering interpretation of the weighted KPCA, hence the name Kernel Spectral Clustering (KSC). Construction of the clustering model together with  
some of its most attractive properties are summarized below. Interested readers can find all details elsewhere \cite{alzate&suykens2006_originalwkpca, alzate&suykens2010}.
\subsubsection{Cluster membership encoding-decoding}
\label{sec:cluster-membership-encoding-decoding}
\noindent
The special properties of some selected eigenvectors of the $D^{-1}M_{D}\Omega$ matrix and the corresponding score variables are discussed in detail in \cite{alzate&suykens2006_originalwkpca, alzate&suykens2010}
under the assumption, that the input data set contains $\mathcal{K}$ clusters. It has been shown, that the more similar a subset of the input data are the more collinear their $\mathcal{K}-1$ dimensional representations 
in the subspace spanned by the columns of the $Z=[\bm{z}^{(1)},\dots,\bm{z}^{(\mathcal{K}-1)}] \in \mathbb{R}^{N\times\mathcal{K}-1}$ score matrix that corresponds to the $\mathcal{K}-1$ leading eigenvectors
of the $D^{-1}M_{D}\Omega$ matrix. Moreover, well separated clusters are mapped into different orthant of this $\mathcal{K}-1$ space. These makes possible the \textit{cluster membership encoding} 
and model construction either by selecting the $\mathcal{K}$ most frequent $\mathcal{K}-1$ dimensional 
sign based code words \cite{alzate&suykens2006_originalwkpca, alzate&suykens2010}, constructed from the rows of $Z$, 
or finding $\mathcal{K}$ direction based encoding exploiting the above mentioned collinearity \cite{Rocco_softKSC}. 
Either way, the model construction results in $\mathcal{K}$ cluster prototypes that can be used then to assign the individual 
data points to one of the clusters. This is done by selecting the cluster that yields the minimal distance measured between the 
$\mathcal{K}$ cluster prototypes and the $\mathcal{K}-1$ dimensional score space representation of a given data point, i.e. the 
corresponding row of $Z$. The distance is either Hamming or direction based depending on the selected encoding.
\subsubsection{Out-of-sample extension and model selection}
\label{sec:out-of-sample-extension-and-model-selection}
\noindent
The fact that KSC includes the above mentioned model construction step before the cluster membership assignment has important consequences. 

The first is the natural way that KSC offers for \textit{out-of-sample extension}. This is simple because the KSC model can be constructed based on a $\mathcal{D}^{tr}=\{\bm{x}_{i}^{tr}\}_{i=1}^{N_{tr}}\subset \mathcal{D}$ 
subset of the input data set and the constructed model can be used then to assign any $\bm{x} \in \mathbb{R}^{d}$ data point to one of the clusters. The model 
construction requires the $\pmb{\beta}^{(k)} \in \mathbb{R}^{N_{tr}}, \quad k=1,...,\mathcal{K}-1$ leading eigenvectors of the 
corresponding $D^{-1}M_{D}\Omega \in \mathbb{R}^{N_{tr}\times N_{tr}}$ matrix. Then before the assignment, one needs to compute the 
\begin{equation}
\label{eq:scorevar}
 z^{(k)}(\bm{x}) = \varphi(\bm{x})^{T}\bm{w}^{(k)}+b^{(k)}=\sum_{i=1}^{N_{tr}}K(\bm{x},\bm{x}_i)\beta_{i}^{(k)}+b^{(k)}
\end{equation}
$k=1,...,\mathcal{K}-1$ projections of the given $\bm{x}$ data point. 

The second is the KSC \textit{model selection} capability. As mentioned above, the collinearity of the $\mathcal{K}-1$ dimensional score space representation of the input data, assigned to the same cluster, indicates how well the data set is partitioned into $\mathcal{K}$ clusters using the given kernel parameter 
and $\mathcal{K}$ cluster number hyper-parameters. This can be exploited for defining model selection criterion that accounts and measures the related collinearity. One can construct KSC models with different hyper-parameter values and then select the one that yields the maximal 
model selection criterion. Combining this with the out-of-sample extension capability, the model construction can be done based on a \textit{training} subset of the data while a different, \textit{validation} subset can be used for the evaluation of the model selection criteria. 
      

\subsection{Original ICD based sparse KSC}
\label{sec:original-ICD-based-sparse-KSC}
\noindent
As discussed in the previous section, KSC requires to compute the $\pmb{\beta}^{(k)} \in \mathbb{R}^{N}, \quad k=1,...,\mathcal{K}-1$ leading eigenvectors of the $D^{-1}M_{D}\Omega$ matrix, assuming $\mathcal{K}$ clusters in the data. The score variables can be computed then based on these eigenvectors,
as given by (\ref{eq:scorevar}), and used for the KSC model construction 
($\bm{x}\in\mathcal{D}$) as well as for clustering any $\bm{x} \in \mathbb{R}^{d}$ data point.

While the \textit{out-of-sample extension} capability of KSC offers the $\mathcal{D}^{tr} \subset \mathcal{D}, N_{tr} = |\mathcal{D}^{tr}|$ based model 
construction, one still would like to use as large $\mathcal{D}^{tr}$ subset as possible in order to incorporate as much information in the model construction as 
available in the entire $\mathcal{D}$ data set. On the other hand, the size of the related eigenvalue problem (\ref{eq:multiway_eigenproblem}) 
grows rapidly with increasing $N_{tr}$ which leads to an intractable problem again in case of large $\mathcal{D}^{tr}$. This requires an approximate solution 
of the associated eigenvalue problem that is suitable even in case of large $N_{tr}$.

Furthermore, the primal solutions of the underlying weighted KPCA problem are expressed as linear combinations of the mapped input data
$\bm{w}^{(k)}=\Phi^{T}\pmb{\beta}^{(k)},k=1,\dotsc,\mathcal{K}-1$. Since the components of the $\pmb{\beta}^{(k)}$
eigenvectors are usually not zero, each data point contributes to the primal variable $\bm{w}^{(k)}$
resulting in a non-sparse model. The so-called reduced set method \cite{Burges_simpliSVdecision, Scholkopf_inputVsFeature} was utilised in  \cite{alzate&suykens_KSC_ICD, Alzate&Suykens_sparseKSC_models2011} to construct the sparse model by finding  
$\mathcal{R}=\{\tilde{\bm{x}}_r\}_{r=1}^{R}\subset \mathcal{D}^{tr}, R<N_{tr}$ reduced set points and the corresponding 
$\bm{\xi}^{(k)}\in \mathbb{R}^{R},k=1,\dotsc,\mathcal{K}-1$ reduced set coefficients 
such that $\bm{w}^{(k)} = \Phi^T\pmb{\beta}^{(k)} \approx \tilde{\bm{w}}^{(k)} =\Psi^{T}\bm{\xi}^{(k)}, k=1,\dotsc,\mathcal{K}-1$
with $\Psi=[\varphi{(\tilde{\bm{x}}_1)},\dotsc,\varphi{(\tilde{\bm{x}}_R)}]^{T}\in \mathbb{R}^{R\times n_{h}}$. 
After finding an appropriate set of reduced set points, the reduced set coefficients are determined by minimizing the squared distance 
of the approximation that yields the 
\begin{equation}
\label{eq:reduced_set_coefs}
 \Omega_{\Psi\Psi}\bm{\xi}^{(k)}=\Omega_{\Psi\Phi}\tilde{\pmb{\beta}}^{(k)}
\end{equation}
linear system at the first order optimality
where $\Omega_{\Phi\Phi}=\Phi\Phi^{T} \in \mathbb{R}^{N_{tr}\times N_{tr}}$,
$\Omega_{\Psi\Phi}=\Psi\Phi^{T} \in \mathbb{R}^{R\times N_{tr}}$ and
$\Omega_{\Psi\Psi}=\Psi\Psi^{T} \in \mathbb{R}^{R\times R}$ are the corresponding kernel matrices.  

Note, that $\bm{w}^{(k)} \approx \tilde{\bm{w}}^{(k)}$ above is nothing more than an approximation based on the linear combination of a small 
$\{\varphi{(\tilde{\bm{x}}_r)}\}_{r=1}^{R}$ subset of the $\{\varphi{(\bm{x}}_i)\}_{i=1}^{N_{tr}}$ feature map vectors. Therefore, finding an appropriate 
reduced set corresponds to identify the $\mathcal{R}=\{\tilde{\bm{x}}_r\}_{r=1}^{R}\subset \mathcal{D}^{tr}, R<N_{tr}$ points such that the 
corresponding $\{\varphi{(\tilde{\bm{x}}_r)}\}_{r=1}^{R}$ feature map vectors are linearly independent.

The incomplete Cholesky decomposition of the $\Omega_{\Phi\Phi}=\Phi\Phi^{T} \in \mathbb{R}^{N_{tr}\times N_{tr}}$ kernel matrix provides solutions to 
both problems mentioned above. It offers a reduced size, approximate solution of the related eigenvalue problem (\ref{eq:multiway_eigenproblem}) 
that is suitable even in case of large $N_{tr}$. Moreover, it automatically provides an appropriate reduced set 
$\mathcal{R}=\{\tilde{\bm{x}}_r\}_{r=1}^{R}\subset \mathcal{D}^{tr}, R<N_{tr}$ that can be used for the sparse KSC model construction as discussed above. 

\subsubsection{On the ICD of the kernel matrix}
\noindent
Any symmetric positive definite matrix $A\in \mathbb{R}^{M\times M}$ can be decomposed as $A=LL^{T}$ where
$L \in \mathbb{R}^{M\times M}$ is a lower triangular matrix. If the spectrum of $A$ decays rapidly it has
a small numerical rank \cite{Williams&Seeger2000} and $A$ can be well approximated by $GG^{T}$ where
$G\in \mathbb{R}^{M\times R}, R\ll M$ \cite{Wright_1999_ICD}. This is the incomplete Cholesky decomposition
of $A$. ICD with symmetric pivoting greedily selects the columns of $A$ and
calculates the columns of $G$ such that a lower bound on the actual gain in the corresponding approximation error 
$\|PAP^{T}-GG^{T}\|_{1}=Tr(PAP^{T}-GG^{T}) = \epsilon$ is maximised \cite{Bach&Jordan_lowrank_2005}.
$P$ is the permutation matrix associated to the symmetric pivoting and $\|\cdot\|_{1}$ is the trace norm. The algorithm 
terminates when the approximation error drops below a certain limit $\epsilon \leq \epsilon_{tol}$. 

Since the spectrum of the $\Omega_{\Phi\Phi}=\Phi\Phi^{T} \in \mathbb{R}^{N_{tr}\times N_{tr}}$  kernel matrix decays rapidly in 
case of many kernels, it can be very often well approximated by 
$GG^T = \tilde{\Omega}_{\Phi\Phi} \approx \Omega_{\Phi\Phi},\; G \in \mathbb{R}^{N_{tr}\times R}$ with a low $R\ll N_{tr}$ rank
\cite{Williams&Seeger2000,smola2000sparse,Bach&Jordan_ICA2002}.
This is utilised in \cite{alzate&suykens_KSC_ICD, Alzate&Suykens_sparseKSC_models2011} 
to obtain an approximate, reduced size solution of the eigenvalue problem given by (\ref{eq:multiway_eigenproblem}), that is suitable even in case of large
$N_{tr}$.

Moreover, as the Cholesky decomposition of the $\Omega_{\Phi\Phi}$ kernel matrix is equivalent to the QR factorization of the corresponding feature map 
$\Phi = [\varphi(\bm{x}_{1}), \dots, \varphi(\bm{x}_{N_{tr}})  ]^T \in \mathbb{R}^{N_{tr}\times n_{h}}$
\cite{shawe2004kernel}, its incomplete version $GG^T = \tilde{\Omega}_{\Phi\Phi} \approx \Omega_{\Phi\Phi}$ 
can be seen as the result of a truncated, pivoted Gram-Schmidt(GS) orthogonalization of these feature map vectors.  
The selected pivots, i.e. $\{\varphi(\tilde{\bm{x}}_r)\}_{r=1}^{R}$ that correspond to the columns of $\Omega$ selected during the ICD,
are linearly independent. Therefore, ICD automatically provides a suitable reduced set $\mathcal{R}=\{\tilde{\bm{x}}_r\}_{r=1}^{R}$ 
such that the corresponding $\{\varphi(\tilde{\bm{x}}_r)\}_{r=1}^{R}$ linearly independent feature vectors are suitable for the sparse KSC 
model construction. 

\subsubsection{The original algorithm}
\noindent
As mentioned above, the $GG^T = \tilde{\Omega}_{\Phi\Phi} \approx \Omega_{\Phi\Phi},\; G \in \mathbb{R}^{N_{tr}\times R}$ ICD of the 
$\mathcal{D}^{tr}$ training data $\Omega_{\Phi\Phi}$ kernel matrix is exploited in two ways in the original sparse KSC algorithm \cite{alzate&suykens_KSC_ICD,Alzate&Suykens_sparseKSC_models2011}. 

First, the size of the related eigenvalue problem (\ref{eq:multiway_eigenproblem}) is reduced from $N_{tr}$ to $R\ll N_{tr}$ by using the 
the $G=U\Lambda V^{T}$ Singular Value Decomposition (SVD) in the corresponding low rank approximation of the training data kernel matrix
$\Omega_{\Phi\Phi} \approx \tilde{\Omega}_{\Phi\Phi} = GG^T = U\Lambda^2U^T$  in (\ref{eq:multiway_eigenproblem}). This leads to the 
\begin{equation}
\label{eq:org_eigenproblem}
 U^{T}\tilde{D}^{-1}M_{\tilde{D}}U\Lambda^{2}\pmb{\gamma}^{(k)}=\tilde{\lambda}^{(k)}\pmb{\gamma}^{(k)},
    \quad k=1,...,\mathcal{K}
\end{equation}
eigenvalue problem with $\pmb{\gamma}^{(k)}=U^T\pmb{\beta}^{(k)}$ such that the approximated eigenvectors of the original problem 
can be obtained as $\tilde{\pmb{\beta}}^{(k)}=U\pmb{\gamma}^{(k)}$. $\tilde{\cdot}$ denotes approximation of the corresponding 
quantity based on the $\Omega_{\Phi\Phi} \approx \tilde{\Omega}_{\Phi\Phi} = GG^T$ low rank approximation.

After the $\tilde{\pmb{\beta}}^{(k)}$ approximated eigenvectors are determined, the sparse solution of the KPCA can be composed that 
corresponds to the $\bm{w}^{(k)} = \Phi^T\pmb{\beta}^{(k)} \approx \tilde{\bm{w}}^{(k)} =\Psi^{T}\bm{\xi}^{(k)}, k=1,\dotsc,\mathcal{K}-1$
approximation.  The ICD of the training data kernel matrix is utilised again at this point by taking the pivots selected during the decomposition 
as the $\mathcal{R}=\{\tilde{\bm{x}}_r\}_{r=1}^{R}\subset \mathcal{D}^{tr}$ reduced set. The $\bm{\xi}^{(k)}\in \mathbb{R}^{R},k=1,\dotsc,\mathcal{K}-1$ 
reduced set coefficients can then be determined by solving (\ref{eq:reduced_set_coefs}) with the  $\Omega_{\Psi\Psi}$ reduced-reduced, 
$\Omega_{\Psi\Phi}$ reduced-training set kernel matrices and the corresponding $\tilde{\pmb{\beta}}^{(k)}$ approximated eigenvectors.  

The approximated score variables of any $\bm{x}\in\mathbb{R}^{d}$ data can be expressed as 
\begin{equation}
\label{eq:approximated-score-variables}
z^{(k)}(\bm{x}) \approx \tilde{z}^{(k)}(\bm{x}) = \sum_{r=1}^{R}K(\bm{x},\bm{x}_r)\xi_{r}^{(k)}+\tilde{b}^{(k)}
\end{equation}
$k=1,\dotsc,\mathcal{K}-1$ by using the above reduced set based sparse approximation in (\ref{eq:scorevar}). The corresponding sparse clustering model can then be 
constructed based on the approximated score variables related to the training data set and any data point can be assigned to one of the 
underlying clusters as described in sections \ref{sec:cluster-membership-encoding-decoding} and \ref{sec:out-of-sample-extension-and-model-selection}.


\section{Main Results}
\subsection{Modified Algorithm}
\label{sec:mod}

\subsubsection{Making it faster}
\label{sec:mod-speed}
\noindent
The computational bottleneck of the original ICD based sparse KSC algorithm \cite{alzate&suykens_KSC_ICD,Alzate&Suykens_sparseKSC_models2011} 
is the computation of the $\tilde{\pmb{\beta}}^{(k)}$ approximated eigenvectors of the $D^{-1}M_{D}\Omega$ matrix associated to 
the $\mathcal{D}^{tr} \subset \mathcal{D}, N_{tr}=|\mathcal{D}^{tr}|$ training data set. 
As mentioned above, the algorithm first exploits the $GG^T = \tilde{\Omega}_{\Phi\Phi} \approx \Omega_{\Phi\Phi}$ ICD of the training set kernel 
matrix then the $G=U{\Lambda}V^T$ SVD of the resulted $G \in \mathbb{R}^{N_{tr}\times R}$ matrix. The corresponding original 
$N_{tr}\times N_{tr}$ sized eigenvalue problem (\ref{eq:multiway_eigenproblem}) is then relaxed to solve (\ref{eq:org_eigenproblem}).
Indeed, the size of the matrix involved in (\ref{eq:org_eigenproblem}) is $R{\times}R$ with $R\ll N_{tr}$
that makes possible to use significantly larger $N_{tr}$ training data sets. 
However, the computation time, required to obtain the corresponding 
eigenvectors, was reported to increase rapidly with $N_{tr}$ by the authors \cite{alzate&suykens_KSC_ICD}, reaching already to $\sim$hours 
with $N_{tr} \sim 10^{5}$. This prevented the algorithm to gain any practical importance as these eigenvectors need to be computed several times 
during the hyper-parameter tuning.  

Having a closer look to the algorithm, one can recognise that the naive computation of the above mentioned $R{\times}R$ sized matrix, 
involved in (\ref{eq:org_eigenproblem}), can lead to a $\mathcal{O}(R^2N_{tr}^2)$ computational complexity.
This might be avoided by exploiting both the special structure of the individual matrices and the effects of the corresponding operations ensuring to 
maintain a $\mathcal{O}(R^2N_{tr})$ complexity. However, the algorithm still includes the $G=U{\Lambda}V^T$ SVD of the large 
$G \in \mathbb{R}^{N_{tr}\times R}$ matrix that keeps the algorithm computationally demanding with increasing $N_{tr}$.
All these complications can be fully avoided by computing the required $\tilde{\pmb{\beta}}^{(k)}$ approximated eigenvectors of the 
$D^{-1}M_{D}\Omega$ matrix in the following alternative way.

Instead of performing the SVD of $G$ then constructing and solving (\ref{eq:org_eigenproblem}), one can transform (\ref{eq:multiway_eigenproblem}) 
to the corresponding 
\begin{equation}
\label{eq:symmetric_eigenproblem}
  D^{-\frac{1}{2}}M_{D}\Omega M_{D}^{T}D^{-\frac{1}{2}}\boldsymbol{\alpha}^{(k)}=\lambda^{(k)}\boldsymbol{\alpha}^{(k)}
  ,\quad k=1,\dotsc,\mathcal{K}-1
\end{equation}
symmetric problem where $\pmb{\alpha}^{(k)}=D^{1/2}\pmb{\beta}^{(k)},k=1,\dotsc,\mathcal{K}-1$ with the same $\lambda^{(k)}$ eigenvalues
as in (\ref{eq:multiway_eigenproblem}). The $\Omega_{\Phi\Phi} \approx \tilde{\Omega}_{\Phi\Phi} = GG^T$ ICD based approximation of the kernel 
matrix can be exploited at this point leading to 
$D^{-\frac{1}{2}}M_{D}\Omega M_{D}^{T}D^{-\frac{1}{2}}\approx \tilde{D}^{-\frac{1}{2}}M_{\tilde{D}}GG^{T} M_{\tilde{D}}^{T}\tilde{D}^{-\frac{1}{2}}$.
Taking $X=\tilde{D}^{-1/2}M_{\tilde{D}}G\in\mathbb{R}^{N_{tr}\times R}$,
performing its QR factorisation that leads to $X=Q_{X}R_{X}$ with $Q_{X}\in\mathbb{R}^{N_{tr}\times R}$ and
$R_{X}\in\mathbb{R}^{R\times R}$, then utilising the $R_{X} =U_{R_{X}}\Sigma_{R_{X}}V_{R_{X}}^{T}$ SVD of this small $R_{X}$ matrix result in
the following eigenvalue decomposition 
\begin{equation}
\label{eq:symmetric_eigenproblem_approximate}
     \tilde{D}^{-\frac{1}{2}}M_{\tilde{D}}GG^{T} M_{\tilde{D}}^{T}\tilde{D}^{-\frac{1}{2}} 
       = [Q_{X}U_{R_{X}}]\Sigma_{R_{X}}^{2}[Q_{X}U_{R_{X}}]^{T}
\end{equation} 
with $Q_{X}U_{R_{X}}\in\mathbb{R}^{N_{tr}\times R}$ eigenvectors and
$\Sigma_{R_{X}}^{2}=\Lambda_{R_{X}}\in\mathbb{R}^{R\times R}$ $\diag(\Lambda_{R_{X}})$ eigenvalues of the
$\tilde{D}^{-\frac{1}{2}}M_{\tilde{D}}GG^{T} M_{\tilde{D}}^{T}\tilde{D}^{-\frac{1}{2}}\in\mathbb{R}^{N_{tr}\times N_{tr}}$ matrix.

Therefore, the $\tilde{\pmb{\alpha}}^{(k)},k=1,\dotsc,\mathcal{K}-1$ leading eigenvectors of the
$\tilde{D}^{-\frac{1}{2}}M_{\tilde{D}}GG^{T} M_{\tilde{D}}^{T}\tilde{D}^{-\frac{1}{2}}$
matrix can be obtained by simply taking the $\mathcal{K}-1$ columns of the $Q_{X}U_{R_{X}}$ matrix that correspond to
the $\mathcal{K}-1$ largest eigenvalues in the diagonal $\Sigma_{R_{X}}^{2}=\Lambda_{R_{X}}$. 
Then the $\Omega_{\Phi\Phi} \approx \tilde{\Omega}_{\Phi\Phi} = GG^T$ ICD based approximated
eigenvectors $\tilde{\pmb{\beta}}^{(k)}$ of
the $D^{-1}M_{D}\Omega$ matrix can be calculated easily from the corresponding
$\tilde{\pmb{\alpha}}^{(k)}$ eigenvectors as
$\tilde{\pmb{\beta}}^{(k)}=\tilde{D}^{-1/2}\tilde{\pmb{\alpha}}^{(k)}$.
These modifications were motivated by \cite{frederix2013sparse} where a similar trick was used to accelerate the computation of some selected eigenvectors 
of the symmetric normalised graph Laplacian. 

It must be noted, that this modified and the original computation of the required approximate eigenvectors of $D^{-1}M_{D}\Omega$ matrix 
give identical results. However, the proposed modification has several advantages. First and most importantly, while the original algorithm includes an SVD
of the large $G\in\mathbb{R}^{N_{tr}\times R}$, the proposed algorithm performs only a much simpler QR factorisation on the large
$X=\tilde{D}^{-1/2}M_{\tilde{D}}G\in\mathbb{R}^{N_{tr}\times R}$ and an SVD only on the small $R_{X}\in\mathbb{R}^{R\times R}$
which can be done significantly faster. Furthermore, since $\tilde{D}^{-1/2}$ is diagonal and the effect 
of left multiplying $G$ by $M_{\tilde{D}}$ is that it removes the $1/\tilde{d}_i$ weighted mean from each column of $G$, $X$ can be
generated quickly column-by-column in place of $G$. Since the QR factorisation of $X$ can then also be done in place, the proposed 
algorithm does not need any extra memory. The greatly reduced computation time of the approximate eigenvectors is clearly demonstrated in the next section.

\subsubsection{Improved sparsity}
\label{sec:mod-sparsity}
\noindent
It has been shown in \cite{Alzate&Suykens_sparseKSC_models2011}, that the original ICD based sparse KSC algorithm is not only 
computationally more demanding but also requires remarkably larger reduced set size $R$ compared to its alternative. 

As briefly discussed in Section \ref{sec:cluster-membership-encoding-decoding}, clusters are well separated in the score variable 
space, lying even in different orthant, that eventually makes possible the clustering model construction. 
This separation is due to the weighted centering of the feature map, achieved through the $b^{(k)}$ bias 
term, that leads to a similar centering of the score variables in (\ref{eq:scorevar}) and (\ref{eq:approximated-score-variables}) \cite{alzate&suykens2010}.
Therefore, the accuracy of the $\tilde{b}^{(k)}$ approximated bias term in (\ref{eq:approximated-score-variables}) can greatly influence the quality 
of the constructed clustering model due to the importance of this centering. 

As mentioned in Section (\ref{sec:SC_as_wKPCA}), the KKT optimality conditions provide the expression for the bias term that has the form of 
$b^{(k)}=-\bm{\mathds{1}}_{N}^{T}D^{-1}\Omega_{\Phi\Phi}\pmb{\beta}^{(k)}  /  [\bm{\mathds{1}}_{N}^{T}D^{-1}\bm{\mathds{1}}_{N}]$ in case of KSC. 
In the original ICD based sparse KSC algorithm, the corresponding $b^{(k)}\approx\tilde{b}^{(k)}$ approximate values, 
required in (\ref{eq:approximated-score-variables}), are simple estimated based on the small reduced set, i.e. computing the inverse degrees 
based on the small $\Omega_{\Psi\Psi} \in \mathbb{R}^{R\times R}$ reduced set kernel matrix and relying on 
the $\bm{w}^{(k)} = \Phi^T\pmb{\beta}^{(k)} \approx \Psi^{T}\bm{\xi}^{(k)}$ approximation. Therefore, the accuracy of this estimate depends 
on the $R$ reduced set.

A more accurate value of $\tilde{b}^{(k)}$ can be obtained relying consistently on the $\Omega_{\Phi\Phi}\approx\tilde{\Omega}=GG^T$ ICD based low rank 
approximation of the entire training set kernel matrix instead of only its reduced subset. This leads to the 
$D\approx\tilde{D}$, $\diag{(\tilde{D})}=\tilde{D}\bm{\mathds{1}}_{N_{tr}}=\tilde{\Omega}_{\Phi\Phi}\bm{\mathds{1}}_{N_{tr}}=G[G^T\bm{\mathds{1}}_{N_{tr}}]$, 
$\pmb{\beta}^{(k)}\approx\tilde{\pmb{\beta}}^{(k)}$ approximations and the simpler $\tilde{b}^{(k)}=[\tilde{\lambda}^{(k)}-1][\bm{\mathds{1}}_{N_{tr}}^{T}\tilde{D}]\tilde{\pmb{\beta}^{(k)}}/N_{tr}$ expression, also given by KKT optimality conditions.
Since all required quantities are already calculated during the computation of the $\tilde{\pmb{\beta}}^{(k)}$ approximate eigenvectors, 
including the corresponding $\lambda^{(k)}\approx\tilde{\lambda}^{(k)}$ approximate eigenvalues as the diagonals of 
$\Sigma_{R_{X}}^{2}=\Lambda_{R_{X}}$ in (\ref{eq:symmetric_eigenproblem_approximate}), these approximated bias terms can be computed 
very quickly.

The original algorithm requires a larger $R$ reduced set size just to obtain an estimate of the $\tilde{b}^{(k)}$ bias term 
in (\ref{eq:approximated-score-variables}) that is good enough to provide an 
appropriate centering which is essential for an accurate clustering model construction. In contrast, the proposed modification results 
in more accurate values, depending less directly on $R$, leading to correct centering in the score variable space. Therefore, the modified 
algorithm can lead to accurate clustering model even at lower reduced set sizes increasing significantly the KSC model sparsity 
which is clearly demonstrated in the next section.

\subsection{Computational experiments}

\subsubsection{Implementation and configuration}
\noindent
The modified ICD based sparse KSC algorithm has been implemented in the \texttt{leuven} test environment \cite{libleuven,libleuven-ksc} 
using the C++ object-oriented language. This lightweight framework provides the possibility of utilising the most popular 
optimised BLAS \cite{BLAS_ref} and LAPACK \cite{LAPACK_ref} numerical linear algebra libraries as computing backends. 
While this multi-threaded implementation can also exploit multiple CPU cores and even NVIDIA GPUs through the appropriate CUDA 
libraries \cite{CUDA} when available, all reported experiments were carried out using only a single Intel core i7 CPU core
with the optimised Intel MKL BLAS/LAPACK backend (v-2019.4.223) \cite{mkl} on MacOS (High Sierra 10.13.6) system with 
a Clang (v-1000.11.45.5) compiler. 

The sparse KSC model is constructed during the \textit{training phase} based on $N_{tr}=|\mathcal{D}^{tr}| \leq N$ training data points sampled uniformly 
random from the entire $\mathcal{D}$ data set. The $\mathcal{R}\subset \mathcal{D}^{tr}$ reduced set is formed by the pivots selected during the ICD of 
the training data kernel matrix. The result of this ICD is the input of the training phase, which includes the computation of the $\tilde{\pmb{\beta}}^{(k)}$ 
approximated eigenvectors of the $D_{tr}^{-1}M_{D_{tr}}\Omega$ matrix associated to the training data set, the construction of the sparse representation 
by solving (\ref{eq:reduced_set_coefs}) and it terminates by producing the cluster prototypes based on the selected cluster membership encoding 
as discussed in Section \ref{sec:cluster-membership-encoding-decoding}. This last step relies on the sparse approximation based expression 
of the score variables as given by (\ref{eq:approximated-score-variables}). 

The resulted sparse KSC model can then be used to assign any data points to one of the clusters during 
the so called \textit{test} or \textit{out-of-sample extension phase} as described in \ref{sec:out-of-sample-extension-and-model-selection} relying 
again on (\ref{eq:approximated-score-variables}). 

\subsubsection{The intertwined spiral synthetic data}
\label{sec:exp-spiral}
\begin{figure}[!t]
\centering
\includegraphics[width=1.8in]{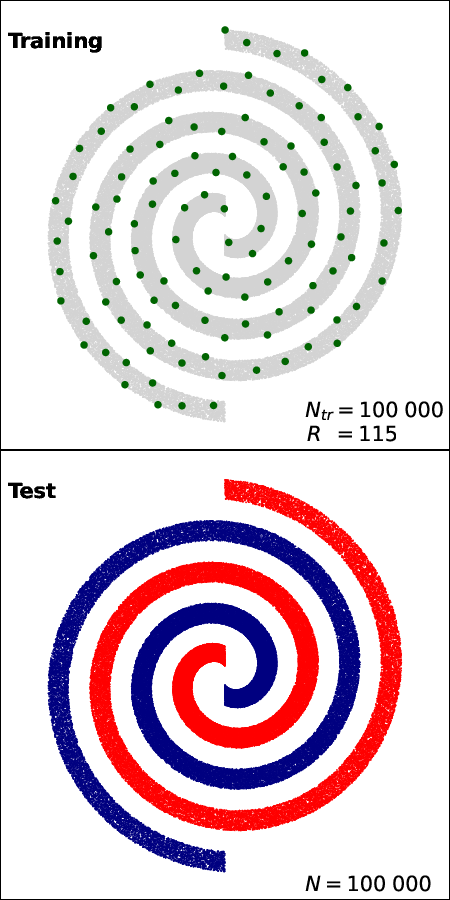}
\caption{ 
 The intertwined spiral synthetic data. \textbf{top}: the entire $N=10^{5}$ input data set is shown
 in grey and the $R=115$ reduced set points, obtained by applying ICD using the entire $N_{tr}=N=10^{5}$ data points with
 $\gamma=0.006$ kernel parameter, are shown in green. \textbf{bottom}: the perfect clustering result, obtained by using
 the proposed algorithm. (Note, that this figure corresponds to the last row of Table \ref{tb:table2}.) 
}
\label{fig:fig1}
\end{figure}

\noindent
The significantly improved computation time is illustrated first by reproducing the same synthetic experiment that was used by the authors of the 
original algorithm to investigate its characteristics and performance in \cite{alzate&suykens_KSC_ICD}. The data set contains 
$N=10^{5}$, $d=2$ dimensional data points, sampled from two intertwined spirals as shown in figure \ref{fig:fig1}. 

\begin{figure}[!t]
\centering
\includegraphics[width=3.5in]{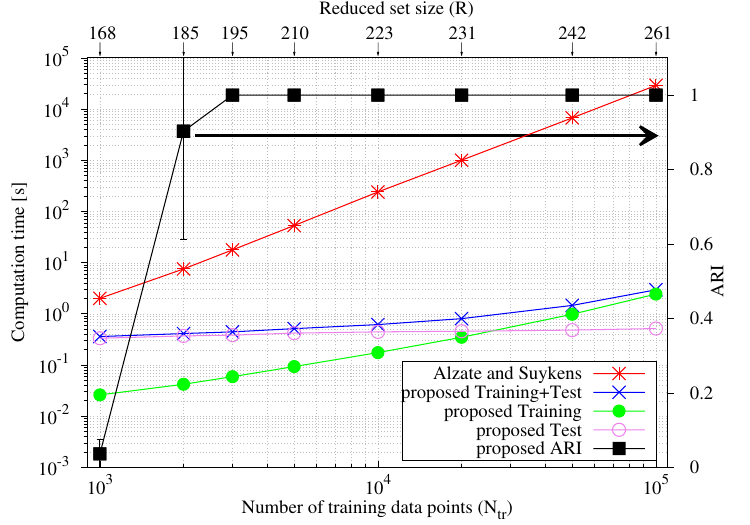}
\caption{Spiral : average computation time and \texttt{ARI} of the proposed algorithm (based on 10 independent runs) as a function of the 
$N_{tr}$ training set size. Numerical values are shown in Table \ref{tb:table1}. The corresponding run times, as reported  by Alzate \& Suykens 
in \cite{alzate&suykens_KSC_ICD}, are also shown for reference.}
\label{fig:fig2}
\end{figure}

The average computation time and Adjusted Rand Index(ARI) \cite{ARI_ref} (based on 10 independent runs), are reported in figure \ref{fig:fig2} as a function 
of the $N_{tr}$ training set size. The corresponding run times, obtained by using the original algorithm in \cite{alzate&suykens_KSC_ICD}, are also shown for reference. 
In order to ensure comparable characteristics and computation times, the reduced set size $R$ was fixed to the same value at each $N_{tr}$ as in 
\cite{alzate&suykens_KSC_ICD}. An RBF kernel was utilised in the form of 
$K(\mathbf{x}_i,\mathbf{x}_j)=\exp(-\|\mathbf{x}_i-\mathbf{x}_j\|_{2}^2/\gamma)$ with kernel parameter of $\gamma_{\text{opt}}=0.006$ that was determined 
during a hyper-parameter tuning by maximising the Balanced Line Fit (BLF) model selection criterion \cite{alzate&suykens2010}. At each different $N_{tr}$ 
training set sizes, the sparse KSC model was constructed during the training phase based on the $N_{tr}$ training data points and the corresponding $R$ reduced 
set size while the entire $N=10^{5}$ data set was partitioned during the test phase. Numerical values are also shown in Table \ref{tb:table1}.
While the original algorithm was reported to require hours to solve this clustering problem when the $N_{tr}$ training set size approaches $10^{5}$ 
\cite{alzate&suykens_KSC_ICD}, it takes only approximately two seconds for the proposed version with the given implementation using exactly 
the same configurations. 

\begin{table}
\begin{center}
\caption{Numerical values of the proposed algorithm shown in figure \ref{fig:fig2}.}
\label{tb:table1}
  \begin{tabular}{| c | c | c | c | c |}
    \hline
    $N_{tr}$ & $R$ & Training [s] & Test [s] & ARI \\
    \hline
    $1\times10^{3}$ & 168 & 0.026 & 0.332 & 0.036$\pm$0.039\\  
    $2\times10^{3}$ & 185 & 0.042 & 0.369 & 0.903$\pm$0.291\\
    $3\times10^{3}$ & 195 & 0.059 & 0.383 & 1.0 \\
    $5\times10^{3}$ & 210 & 0.093 & 0.419 & 1.0 \\
    $1\times10^{4}$ & 223 & 0.174 & 0.442 & 1.0 \\
    $2\times10^{4}$ & 231 & 0.347 & 0.456 & 1.0 \\
    $5\times10^{4}$ & 242 & 0.976 & 0.479 & 1.0 \\
    $1\times10^{5}$ & 261 & 2.428 & 0.514 & 1.0 \\
    \hline
  \end{tabular}
\end{center}
\end{table}

As mentioned above, the $R$ reduced set sizes were fixed to be the same at each $N_{tr}$ as in \cite{alzate&suykens_KSC_ICD} in order to 
ensure comparable characteristics affecting the computing time. The minimum reduced set sizes $R_{\text{min}}$, required to obtain a perfect clustering (average ARI=1), are reported in Table \ref{tb:table2} using the 
\textit{original}($\gamma_{\text{opt}}=0.0102$) and \textit{proposed}($\gamma_{\text{opt}}=0.006$) versions of computing the approximated bias terms in 
(\ref{eq:approximated-score-variables}). As discussed in Section \ref{sec:mod-sparsity}, the accuracy of the bias term approximation influences the quality of the 
corresponding clustering model. Since the related modification leads to a more accurate 
approximation, depending less directly on $R$, the new algorithm requires significantly smaller reduced set sizes for a perfect clustering as 
demonstrated in Table \ref{tb:table2}. Moreover, the related $R_{\text{min}}$ value decreases rapidly between $N_{tr}=10^3-10^4$, then stays constant. This 
indicates that the corresponding clustering model can well exploit and benefit from the new information incorporated into the increasing training data set while 
no significant additional information arrives at $N_{tr}>1-2\times10^4$. 
In contrast, the original version not only requires significantly larger reduced set sizes for a perfect clustering, but the corresponding $R_{\text{min}}$ value
decreases only very mildly with increasing $N_{tr}$. This shows that the related clustering model accuracy is limited more by the corresponding bias term 
approximation as it is determined more directly by the reduced set size.
 

It must be noted, that according to the last row of Table \ref{tb:table2}, the proposed modifications led to a new ICD based sparse KSC algorithm that requires only 
a second to solve the same clustering problem that was reported to take about 8 hours for the original version in \cite{alzate&suykens_KSC_ICD}.  
\begin{table}
\begin{center}
\caption{Spiral: minimum reduced set size $R_{\text{min}}$ and corresponding sparsity in \% required to get a perfect clustering (\texttt{ARI}=1).}
\label{tb:table2}
  \begin{tabular}{| c | c | c | c | c | }
    \hline
                   & \multicolumn{1}{ c |}{Original} & \multicolumn{3}{ c |}{Proposed} \\
    \cline{2-5}
     $N_{tr}$ & $R_{\text{min}}$  & $R_{\text{min}}$  & Training [s] & Test [s]  \\
    \hline
    $3\times10^{3}$ & 270 (91.00\%) & 180 (94.00\%) & 0.052 & 0.356  \\
    $5\times10^{3}$ & 252 (94.96\%) & 138 (97.24\%) & 0.056 & 0.275  \\
    $1\times10^{4}$ & 251 (97.49\%) & 121 (98.79\%) & 0.074 & 0.243  \\
    $2\times10^{4}$ & 246 (98.77\%) & 115 (99.42\%) & 0.155 & 0.232  \\
    $5\times10^{4}$ & 244 (99.51\%) & 115 (99.77\%) & 0.411 & 0.233  \\
    $1\times10^{5}$ & 244 (99.76\%) & 115 (99.88\%) & 0.869 & 0.193  \\
    \hline
  \end{tabular}
\end{center}
\end{table}
\subsubsection{Image segmentation}
\label{sec:exp-image}
\noindent
Ten color images were selected from the Berkeley image data set \cite{MartinFTM01} to demonstrate the performance of the proposed sparse KSC 
algorithm on a real life problem such as image segmentation. 

Each of the RGB color images consists of $321 \times 481$ pixels. A local color histogram was computed at each pixel by taking a $5 \times 5$ window 
around the pixels using minimum variance color quantisation of eight levels. After normalisation, the $N=154\;401$ color histograms serve the $8$ dimensional 
data set of the clustering problem. The $\chi^{2}$ kernel $K(h^{(i)},h^{(j)})=\exp(-\chi_{ij}^{2}/\sigma_{\chi^2})$ was used to compute the similarity between two 
local color histograms $h^{(i)}$ and $h^{(j)}$ with $\sigma_{\chi^2}$ kernel parameter and $\chi_{ij}^{2}= 0.5\sum_{l=1}^{8} (h_{l}^{(i)}-h_{l}^{(j)})^{2} / (h_{l}^{(i)}+h_{l}^{(j)})$ \cite{puzicha1997similaritymeasure,Fowlkes_Nystrom,alzate&suykens2010}.

The so called Balanced Angular Similarity(BAS) \cite{libleuven-ksc}, a direction based cluster membership encoding-decoding similar to the one in 
\cite{Rocco_softKSC}, was utilised in this experiment. The optimal values of the $\sigma_{\chi^2}$ kernel parameter and $\mathcal{K}$ cluster number 
hyper-parameters were determined by maximising the corresponding model selection criterion over a 
$(\mathcal{K}\in\{3,\dotsc,10\},\sigma_{\chi^2}\in [0.001,1.0])$ 2D grid. The previously utilised BLF was also evaluated at 
$\mathcal{K}=2,3$ whenever the optimal cluster number was found to be $\mathcal{K}=3$ (as the BAS based criterion defined only for $\mathcal{K}>2$). 
During the hyper-parameter tuning, a sparse KSC model was constructed at each 2D grid point based on the $N_{tr}=10\;000$ training data points 
while the obtained clustering model was used to partition $N_{v}=20\;000$ independent validation points on which the 
model selection criterion was computed. 

\begin{table}
\begin{center}
\caption{Image Data: details on the ICD and the hyper-parameters.}
\label{tb:table3}
 \begin{tabular}{| r | r  r  r  r | r  r |}
    \hline
    & \multicolumn{4}{ c |}{ICD  (R$_{\text{max}}=500$)}   &  \multicolumn{2}{ c |}{Optimal} \\
    & \multicolumn{4}{ c |}{(input parameters, rank, time)} &  \multicolumn{2}{ c |}{hyper-parameters} \\
    \cline{2-5} \cline{5-7}
    Image ID  &  $\sigma_{\chi^2}$  & $\epsilon_{\text{tol}}$  &  R  &  T[s]  &  $\mathcal{K}$   &   $\sigma_{\chi^2}$  \\
    \hline
    119082 & 0.07 & 0.600 & 196  & 0.260 & 3 & 0.0251  \\
    145086 & 0.05 & 0.600 & 119  & 0.122 & 4 & 0.0841  \\
    147091 & 0.50 & 0.030 & 116  & 0.122 & 2 & 0.0800  \\
    167062 & 0.01 & 0.900 &  98  & 0.085 & 3 & 0.0720  \\
    182053 & 0.20 & 0.010 & 179  & 0.222 & 5 & 0.0022  \\
    196073 & 0.05 & 0.800 & 169  & 0.204 & 2 & 0.0475  \\
    295087 & 0.07 & 0.032 & 196  & 0.253 & 3 & 0.1170  \\
    3096   & 0.01 & 0.800 & 142  & 0.152 & 3 & 0.0660  \\
    42049  & 0.01 & 0.850 & 138  & 0.147 & 3 & 0.0128  \\
    62096  & 0.20 & 0.120 & 192  & 0.246 & 2 & 0.0120  \\
    \hline
  \end{tabular}
\end{center}
\end{table}

The final sparse KSC model was constructed on the training data set using the optimal hyper-parameter values (reported in Table \ref{tb:table3} with some 
details on the ICD phase) and utilised to partition the entire $N=154\;401$ color histograms to produce the image segmentation.  
The corresponding F-measure, with respect to human segmentation, as a performance criterion is reported in Table \ref{tb:table4} together with 
earlier results \cite{alzate&suykens2010} obtained by using the \textit{dense} version of the KSC and the Nystr{\"o}m method. The present  
ICD based sparse KSC algorithm provides better F-measure values than the corresponding dense KSC while it has already been reported in 
\cite{alzate&suykens2010}, that the dense KSC outperforms the Nystr{\"o}m method both in terms of accuracy (F-measure) 
and speed. As discussed in Section \ref{sec:original-ICD-based-sparse-KSC}, the dense KSC algorithm is feasible only for relatively small training data 
sets as it relies on the eigenvalue decomposition of the entire training data kernel matrix. Therefore, the F-measures were achieved in 
\cite{alzate&suykens2010} by constructing a dense KSC model on a small $N_{tr}=2000$ only randomly selected training set then partitioning the 
entire $N=154\;401$ color histograms utilising its out-of-sample extension capability. In contrast, the present sparse KSC model can easily be constructed 
based on significantly larger $N_{tr}=10\;000$ data points exploiting the incorporated information that eventually results in a more accurate clustering model.
Moreover, the proposed modifications make now possible to obtain a more compact, sparse clustering model and to perform all the related computations 
within a fraction of a second demonstrating clearly the benefits of the presented results.
It must be noted, that while the proposed algorithm can handle significantly larger training data sets, it has been found that $N_{tr}>10\;000$ do not provide  
additional information that would yield a significant increase in the segmentation quality.

\begin{table}
\begin{center}
\caption{Image Data: F-measure with respect to human segmentation ($\mathcal{K}, \sigma_{\chi^2}$ hyper-parameters and $R$ reduced set size of 
the proposed ICD based sparse KSC are reported in Table \ref{tb:table3}).}
\label{tb:table4}
 \begin{tabular}{| r | r  r  r | r  r | }
    \hline
    \multicolumn{1}{| c |} {Image}& \multicolumn{3}{ c |}{F-measure} &  \multicolumn{2}{ c |}{Time [s] (proposed)} \\
    \cline{2-4} \cline{5-6} 
    \multicolumn{1}{| c |} {ID}& Nystr{\"o}m & KSC  & Proposed & Training & Test \\
    \hline
    119082 & 0.62          & 0.73          & \textbf{0.77} &   0.190 & 1.08  \\ 
    145086 & 0.78          & \textbf{0.88} & \textbf{0.88} &   0.114 & 0.59  \\ 
    147091 & 0.68          & \textbf{0.80} & \textbf{0.80} &   0.100 & 0.58  \\ 
    167062 & 0.46          & 0.85          & \textbf{0.89} &   0.048 & 0.48  \\ 
    182053 & \textbf{0.71} & 0.65          & \textbf{0.71} &   0.078 & 0.93  \\ 
    196073 & 0.74          & \textbf{0.79} &          0.75 &   0.143 & 0.97  \\ 
    295087 & 0.60          & 0.72          & \textbf{0.74} &   0.174 & 1.03  \\ 
    3096   & 0.27          & 0.72          & \textbf{0.78} &   0.068 & 0.67  \\ 
    42049  & 0.87          & 0.88          & \textbf{0.90} &   0.105 & 0.68  \\ 
    62096  & 0.76          & 0.78          & \textbf{0.82} &   0.084 & 1.00  \\ 
    \hline
  \end{tabular}
\end{center}
\end{table}


\section{Conclusion}
An improved version of the sparse multiway Kernel Spectral Clustering (KSC) algorithm,
exploiting the incomplete Cholesky decomposition (ICD) based low rank approximation of
the kernel matrix, is presented. The computational characteristics of the original algorithm is 
drastically improved, especially when applied on large scale data, by replacing the computationally 
most demanding core part with a significantly faster but equivalent alternative. Furthermore, the 
level of sparsity is also increased significantly by changing the related part of the original algorithm to 
a new, more accurate approximation leading to more compact clustering model representation. 
These result in solving clustering problems now within a second that were reported to require hours 
for the original version with an even more compact clustering model representation with increased descriptive power. 
This transforms the original, only theoretically relevant ICD based sparse KSC algorithm to applicable for large scale 
data clustering problems. The theoretical results are demonstrated by computational experiments on carefully 
chosen synthetic data as well as on real life problems such as image segmentation.

\section*{Acknowledgments}
\noindent
Johan Suykens acknowledges ERC Advanced Grant E-DUALITY (787960), iBOF/23/064, Flemish Government (AI Research Program), Leuven.AI Institute.

\bibliographystyle{apalike}
\bibliography{ref}

\end{document}